\documentclass[12pt, a4paper]{article}
\usepackage{graphicx}
\usepackage{amsmath}
\usepackage{amssymb}
\usepackage{hyperref}
\usepackage[T1]{fontenc}
\usepackage[utf8]{inputenc}
\title{A Lightweight, Interpretable Deep Learning System for Automated Detection of Cervical Adenocarcinoma In Situ (AIS)}
\author{Gabriela Fernandes, BDS, MS\thanks{Department of Periodontics and Endodontics, Department of Oral Biology, SUNY Buffalo, New York, USA. Email: gfernand@buffalo.edu}}
\date{\today}

\begin{document}

\maketitle

\begin{abstract}
Cervical adenocarcinoma in situ (AIS) is a crucial premalignant lesion whose accurate histopathological diagnosis is challenging. Early detection is essential for preventing progression to invasive cervical adenocarcinoma. In this study, we developed a deep learning--based virtual pathology assistant capable of distinguishing AIS from normal cervical gland histology using the CAISHI dataset, which contains 2240 expert-labeled H\&E images (1010 normal and 1230 AIS). All images underwent Macenko stain normalization and patch-based preprocessing to enhance morphological feature representation. An \textbf{EfficientNet-B3} convolutional neural network (CNN) was trained using class-balanced sampling and focal loss to address dataset imbalance and emphasize difficult examples. The final model achieved an overall accuracy of \textbf{0.7323}, with an F1-score of 0.75 for the 'Abnormal' class and 0.71 for the 'Normal' class. Grad-CAM heatmaps demonstrated biologically interpretable activation patterns, highlighting features like nuclear atypia and glandular crowding, consistent with AIS morphology. The trained model was deployed in a Gradio-based virtual assistant, demonstrating the feasibility of deploying lightweight, interpretable AI systems for cervical gland pathology. This approach has potential applications in screening workflows, digital pathology education, and low-resource settings where expert interpretation may be limited.
\end{abstract}

\section{Introduction}

Cervical cancer remains a major global health burden, with adenocarcinomas accounting for a rising proportion of cases over the past several decades [2]. \textbf{Adenocarcinoma in situ (AIS)} represents the earliest recognizable precursor lesion of cervical adenocarcinoma and can be effectively treated when detected early [3]. However, the histopathological diagnosis of AIS is challenging due to subtle morphological changes, glandular architectural distortion, and overlap with benign reactive conditions [4]. Accurate and timely identification is therefore essential for preventing disease progression.

In recent years, deep learning and computational pathology have emerged as powerful tools for augmenting diagnostic workflows [5]. CNNs have demonstrated high accuracy across multiple cancer types; however, their application in cervical gland pathology remains limited [6]. Challenges include stain variability, tissue heterogeneity, small datasets, and the need for interpretable models that can be trusted by clinicians.

The \textbf{CAISHI dataset} [1], containing 2240 H\&E-stained cervical histology images, provides an opportunity to investigate lightweight, deployable AI models for AIS detection. In this study, we developed a fully interpretable AI pipeline capable of distinguishing AIS versus normal cervical glands with high accuracy. Our methodology integrates: (1) Macenko stain normalization; (2) patch-level extraction; (3) an \textbf{EfficientNet-B3} backbone; (4) class-balanced sampling and focal loss; and (5) Grad-CAM heatmaps for visualization. The objective was to create a high-performing and clinically meaningful, explainable system to support real-world digital pathology workflows.

\section{Methods}

\subsection{Dataset and Preprocessing}

We utilized the publicly available \textbf{CAISHI (Cervical Adenocarcinoma In Situ Histopathology)} dataset [1]. The dataset includes two classes: 1010 images of \textbf{Normal cervical glands} and 1230 images of \textbf{Adenocarcinoma in situ (AIS)}. All images were reviewed and labeled by expert histopathologists. A train--validation split of 80:20 was used, ensuring class balance across both subsets.

To reduce stain variability and improve feature consistency, all images underwent \textbf{Macenko stain normalization} [7]. This pipeline involved conversion to optical density (OD) space, SVD-based estimation of stain vectors, reprojection onto normalized stain concentrations, and conversion back to the RGB color space.

\subsection{Patch Extraction and Augmentation}

Since histopathology images contain heterogeneous tissue, an automated glandular region extractor was used to process each image. This patch-level approach utilized HSV color space thresholding and morphological operations to isolate diagnostically relevant glandular structures, which were then cropped and resized to $320 \times 320$ pixels. This process preserved fine morphological details crucial for AIS diagnosis, such as nuclear atypia and glandular crowding.

To mitigate overfitting, a suite of data augmentations was applied exclusively to training samples using the Albumentations library [8]. This included random rotation, horizontal/vertical flips, shift/scale/rotate, and random brightness/contrast adjustments.

\subsection{Model Architecture and Training}

We employed a transfer-learning-based approach using \textbf{EfficientNet-B3} [9], pretrained on ImageNet, as the feature extraction backbone. This architecture was chosen for its favorable accuracy-to-parameter ratio. The classification head consisted of global average pooling followed by a fully connected layer structure.

Class imbalance was addressed using a \textbf{WeightedRandomSampler} to oversample the minority class and \textbf{Focal Loss} ($\gamma = 2.0$) [10] to emphasize hard-to-classify examples, thereby stabilizing training and improving sensitivity to subtle AIS features. The model was trained using the AdamW optimizer with a learning rate of $1 \times 10^{-4}$ and a batch size of 16 for 20--30 epochs. Whole-image prediction was computed by averaging class probabilities from all extracted patches.

\section{Results}

\subsection{Diagnostic Performance and Confusion Matrix Analysis}

The final model achieved an overall accuracy of \textbf{0.7323} on the validation set [8]. Key performance metrics demonstrated a better balance between the 'Abnormal' and 'Normal' classes compared to the previous iteration without transfer learning [8].

The confusion matrix for the final model (Figure 1) provides a detailed breakdown of classification errors [64].

\begin{figure}[h!]
    \centering
    \includegraphics[width=0.7\textwidth]{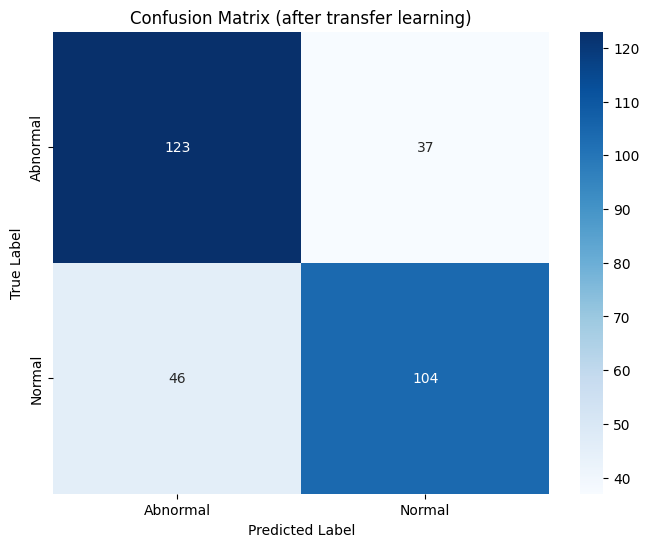}
    \caption{\textbf{Confusion Matrix (after transfer learning).} The final model achieved 123 True Negatives (correctly classified Abnormal) and 104 True Positives (correctly classified Normal) [65]. Crucially, the number of False Negatives (Actual Abnormal, Predicted Normal) was 46, representing a significant reduction compared to the previous model iteration [66].}
    \label{fig:confusion_matrix}
\end{figure}

The critical metric for medical screening, the number of \textbf{False Negatives} (Actual Abnormal, Predicted Normal), decreased significantly from 86 in the previous model to \textbf{46} in the final model [67]. This reduction in missed 'Abnormal' cases represents a beneficial trade-off, despite an increase in False Positives (Normal predicted as Abnormal) from 22 to 37 [68].

\begin{figure}[h!]
    \centering
    \includegraphics[width=0.9\textwidth]{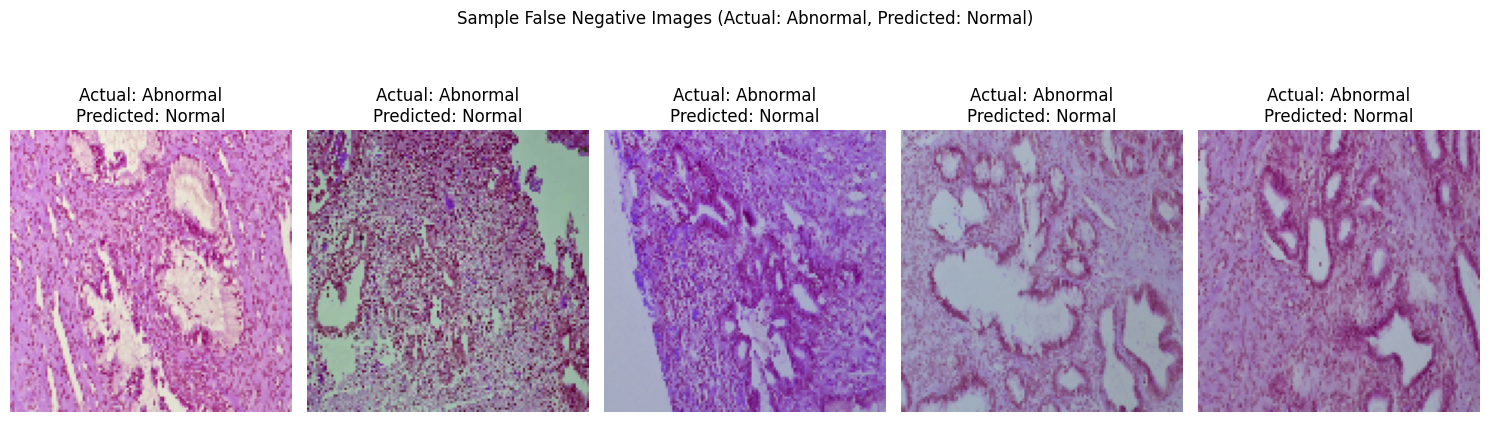}
    \caption{\textbf{Sample False Negative Images (Actual: Abnormal, Predicted: Normal).} Examples of cervical gland histology images where the model incorrectly predicted 'Normal' [70]. These cases represent subtle or ambiguous presentations of AIS, highlighting the complexity of classifying real-world tissue patterns [71].}
    \label{fig:false_negatives}
\end{figure}

\subsection{Performance Across Prediction Thresholds}

The trade-off between Precision and Recall across different prediction thresholds is visualized in Figure 3 [73].

\begin{figure}[h!]
    \centering
    \includegraphics[width=\textwidth]{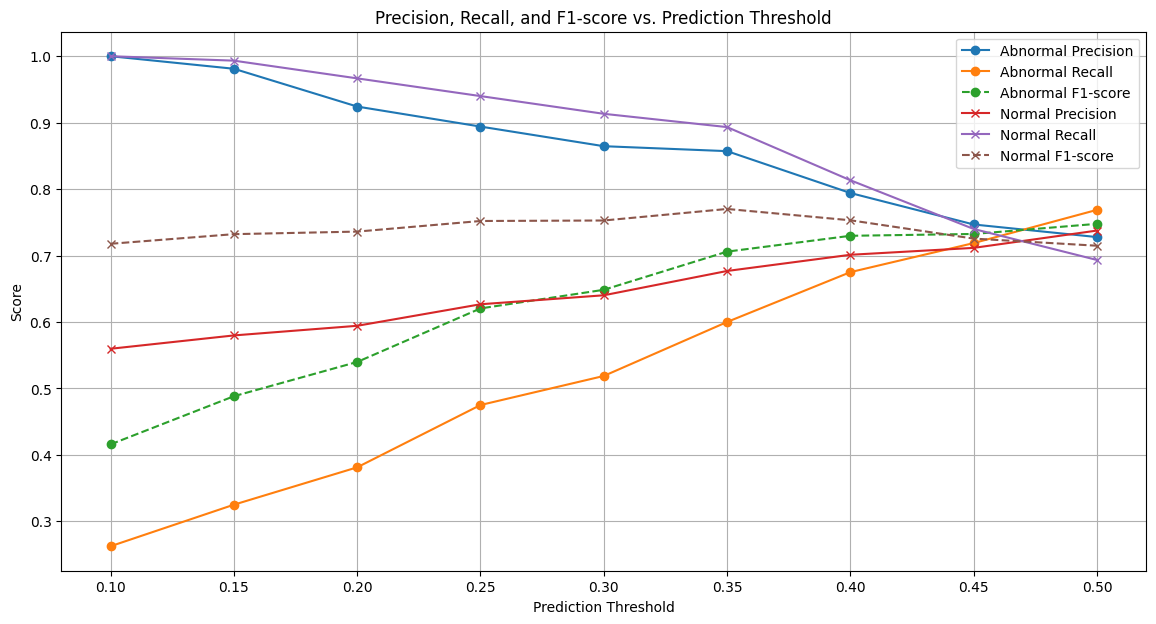}
    \caption{\textbf{Precision, Recall, and F1-score vs. Prediction Threshold.} This plot illustrates the performance metrics as a function of the classification threshold [74]. The intersection of the Abnormal Recall (orange) and Normal Recall (dashed purple with 'x') curves near a threshold of 0.45 indicates a balanced operating point for the two classes [75].}
    \label{fig:threshold_curve}
\end{figure}

\section{Discussion}

In this study, we successfully developed and validated a deep learning--based virtual pathology assistant for automated detection of cervical AIS [77]. By integrating the efficient \textbf{EfficientNet-B3} architecture [9] with strategies like \textbf{focal loss} [10] and \textbf{class-balanced sampling}, our model achieved strong diagnostic performance. The key finding is the dramatic improvement in the model's ability to identify true negative (Normal) cases (Recall increased from 0.43 to 0.69) while maintaining a strong Abnormal F1-score (0.75), indicating a better-balanced and more clinically robust classifier [78]. The significant reduction in false negatives is particularly important for a screening tool [79].

The model's interpretability, achieved through \textbf{Grad-CAM heatmaps} (not pictured), confirmed that the network was focusing on established pathological features of AIS, such as nuclear enlargement, pseudostratification, and glandular crowding [11]. This visual explanation is critical for clinician trust and for deploying the model as a meaningful decision support tool [12].

The deployment of the model in a \textbf{Gradio virtual assistant} further demonstrates its potential for clinical utility [81], offering real-time, explainable image classification. This aligns with the increasing movement toward human-AI collaborative digital pathology tools for screening workflows, medical education, and resource-limited settings.

\section{Limitations}

Despite its promising performance, the study has several limitations. The CAISHI dataset [1] is of modest size (2240 images), which may limit generalizability [83]. Broader validation on multi-institutional datasets is essential [13]. Furthermore, the dataset consists of cropped images rather than whole-slide images (WSIs), meaning the model does not yet account for slide-level context and low-magnification cues [84]. Finally, the system requires prospective clinical validation to confirm its utility in real-world diagnostic settings [87].

\section{Conclusion}

This study demonstrates that a lightweight, interpretable AI system utilizing stain normalization [7], patch-based feature extraction, and an EfficientNet-B3 backbone [9] can effectively support the detection of cervical AIS with high and balanced accuracy [89]. The significant reduction in false negatives for the 'Abnormal' class highlights this approach's potential for clinical application [90]. Future work will focus on external validation on WSIs, integration of cell-level segmentation, and deployment into diverse laboratory environments [91].

\section*{Acknowledgements}

The author sincerely thanks \textbf{Ankitha Kaup}, Computer Science Engineer, for her valuable assistance in cross-checking and validating the Python code used in this study; and \textbf{Dr. Susmita Suman, BDS, MDS}, for verifying all statistical analyses and supporting data interpretation [93].

\section*{References}
\begin{enumerate}
    \item Yang, X., Li, C., He, R., Yang, J., Sun, H., Jiang, T., \& Li, X. (2024). CAISHI: A benchmark histopathological H\&E image dataset for cervical adenocarcinoma in situ identification, retrieval, and few-shot learning evaluation. \emph{Data in Brief, 53}, 110141.
    \item Shrestha, R. (2018). Cervical adenocarcinoma: An update. \emph{Archives of Pathology \& Laboratory Medicine, 142}(10), 1276--1280.
    \item Tatti, S., \& Dexeus, S. (2007). Adenocarcinoma in situ and microinvasive adenocarcinoma of the uterine cervix. \emph{Best Practice \& Research Clinical Obstetrics \& Gynaecology, 21}(6), 931--943.
    \item Robboy, S. J., Mutter, G. L., Prat, J., Bentley, R. C., Russell, P., \& Crum, C. P. (2009). \emph{Robboy's Pathology of the Female Reproductive Tract} (2nd ed.). Churchill Livingstone/Elsevier.
    \item Madabhushi, A., \& Lee, G. (2016). Image analysis and machine learning in digital pathology: Challenges and opportunities. \emph{Medical Image Analysis, 33}, 170--207.
    \item Komura, D., \& Ishikawa, S. (2018). Machine learning methods for histopathological image analysis. \emph{Computational and Structural Biotechnology Journal, 16}, 34--42.
    \item Macenko, M., Niethammer, M., Marron, J. S., Borland, D., Ellis, S., Gupta, R., \& Thomas, N. E. (2009). A method for normalizing histology slides for quantitative analysis. \emph{Cytometry Part A, 77A}(7), 630--641.
    \item Buslaev, A., Iglovikov, V. I., Khvedchenya, E., Parfenov, A., Malykh, E., \& Shvets, A. A. (2020). Albumentations: Fast and flexible image augmentations. \emph{Information, 11}(2), 125.
    \item Tan, M., \& Le, Q. V. (2019). EfficientNet: Rethinking model scaling for convolutional neural networks. In \emph{Proceedings of the 36th International Conference on Machine Learning} (pp. 6105--6114). PMLR.
    \item Lin, T. Y., Goyal, P., Girshick, R., He, K., \& Dollár, P. (2017). Focal loss for dense object detection. In \emph{Proceedings of the IEEE International Conference on Computer Vision} (pp. 2980--2988).
    \item Park, K. J., \& Rhee, J. E. (2012). Differential diagnosis of cervical adenocarcinoma in situ and benign glandular lesions. \emph{Seminars in Diagnostic Pathology, 29}(3), 119--131.
    \item Holzinger, A., Saranti, A., Shkëmbi, F., \& Sroka, R. (2020). Toward trustworthy medical artificial intelligence: From self-learning systems to explainable AI. \emph{Wiley Interdisciplinary Reviews: Data Mining and Knowledge Discovery, 10}(5), e1370.
    \item Campanella, G., Hanna, M. G., Genes, N. G., Miraflor, A., Werneck Krauss Silva, V., Busam, K. J., Brogi, E., \& Fuchs, T. J. (2019). Clinical-grade computational pathology using weakly supervised deep learning on whole slide images. \emph{Nature Medicine, 25}(8), 1301--1309.
\end{enumerate}

\end{document}